\documentclass[10pt,twocolumn,letterpaper]{article}

\usepackage{cvpr}              

\definecolor{cvprblue}{rgb}{0.21,0.49,0.74}
\usepackage[pagebackref,breaklinks,colorlinks,allcolors=cvprblue]{hyperref}
\usepackage{mathtools}
\usepackage{amsmath}
\usepackage{multirow}
\usepackage{wrapfig}
\usepackage{amssymb}
\usepackage{pifont}
\usepackage{threeparttable}
\usepackage{booktabs}

\usepackage{color,colortbl}
\definecolor{ggray}{rgb}{0.90, 0.90, 0.98}

\def\paperID{4210} 
\def\confName{CVPR}
\def\confYear{2025}

\title{SegMAN: Omni-scale Context Modeling with State Space Models \\ and Local Attention for Semantic Segmentation}

\newcommand*\samethanks[1][\value{footnote}]{\footnotemark[#1]}

\author{%
  Yunxiang Fu\thanks{Equal contribution}
  \ \ \ \
  Meng Lou\samethanks 
  \ \ \ \
  Yizhou Yu \\
  School of Computing and Data Science, The University of Hong Kong
  	\\
	{\tt \small yunxiang@connect.hku.hk}, {\tt \small loumeng@connect.hku.hk},  {\tt \small yizhouy@acm.org}
}

\begin{document}
\maketitle
\begin{abstract}
High-quality semantic segmentation relies on three key capabilities: global context modeling, local detail encoding, and multi-scale feature extraction. However, recent methods struggle to possess all these capabilities simultaneously. Hence, we aim to empower segmentation networks to simultaneously carry out efficient global context modeling, high-quality local detail encoding, and rich multi-scale feature representation for varying input resolutions. In this paper, we introduce SegMAN, a novel linear-time model comprising a hybrid feature encoder dubbed SegMAN Encoder, and a decoder based on state space models. Specifically, the SegMAN Encoder synergistically integrates sliding local attention with dynamic state space models, enabling highly efficient global context modeling while preserving fine-grained local details. Meanwhile, the MMSCopE module in our decoder enhances multi-scale context feature extraction and adaptively scales with the input resolution. Our SegMAN-B Encoder achieves 85.1\% ImageNet-1k accuracy (+1.5\% over VMamba-S with fewer parameters). When paired with our decoder, the full SegMAN-B model achieves 52.6\% mIoU on ADE20K (+1.6\% over SegNeXt-L with 15\% fewer GFLOPs), 83.8\% mIoU on Cityscapes (+2.1\% over SegFormer-B3 with half the GFLOPs), and 1.6\% higher mIoU than VWFormer-B3 on COCO-Stuff with lower GFLOPs. Our code is available at \url{https://github.com/yunxiangfu2001/SegMAN}.
\end{abstract}

\section{Introduction}
\label{sec:intro}

\begin{figure}[t]
\centering
\includegraphics[width=0.98\columnwidth]{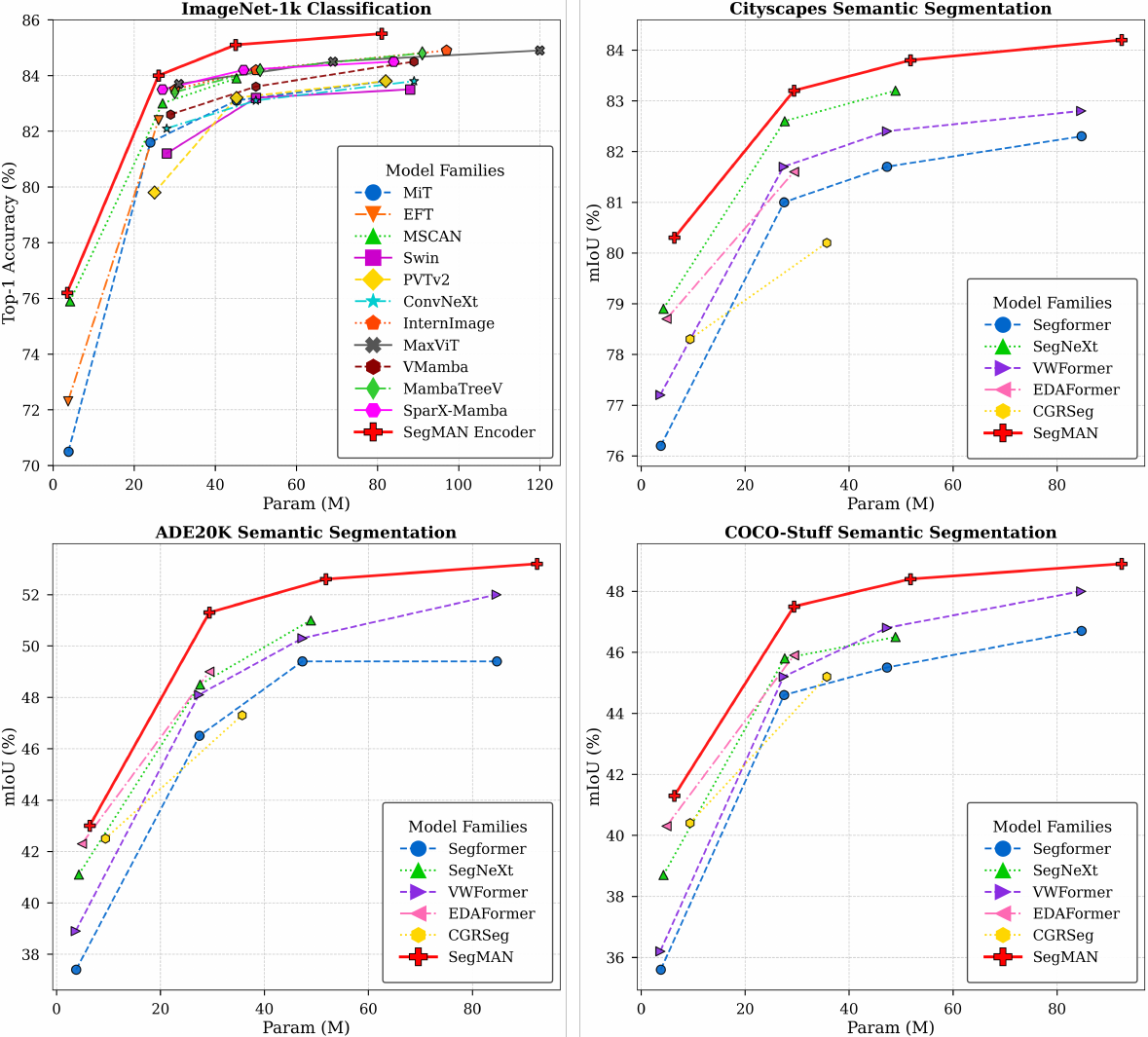}
\caption{SegMAN Encoder classification performance compared with representative vision backbones alongside semantic segmentation results of the full SegMAN model compared to prior state-of-the-art models.}
\label{segman_performance}
\vspace{-5mm}
\end{figure}

\begin{figure*}[t]
\centering
    \begin{minipage}[t]{0.3\textwidth}
\centering
\makeatletter\def\@captype{figure}
  \setlength{\abovecaptionskip}{0.1cm}
  \setlength{\belowcaptionskip}{0.2cm}
\includegraphics[width=\textwidth]{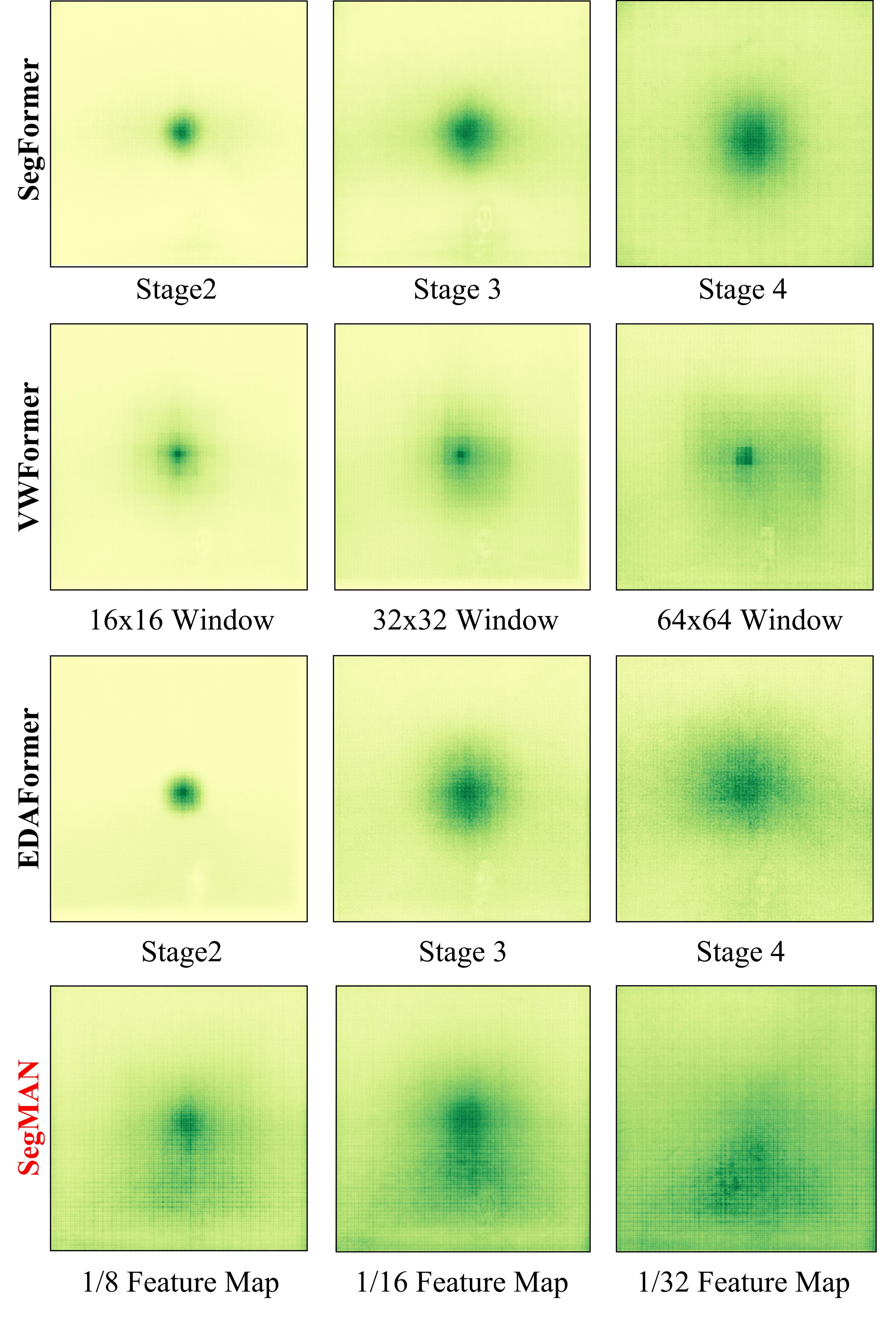}
\label{scaling}
\end{minipage}
\quad
\begin{minipage}[t]{0.55\textwidth}
\centering
\makeatletter\def\@captype{figure}
\setlength{\abovecaptionskip}{0.1cm}
\setlength{\belowcaptionskip}{0.2cm}
    \includegraphics[width=\textwidth]{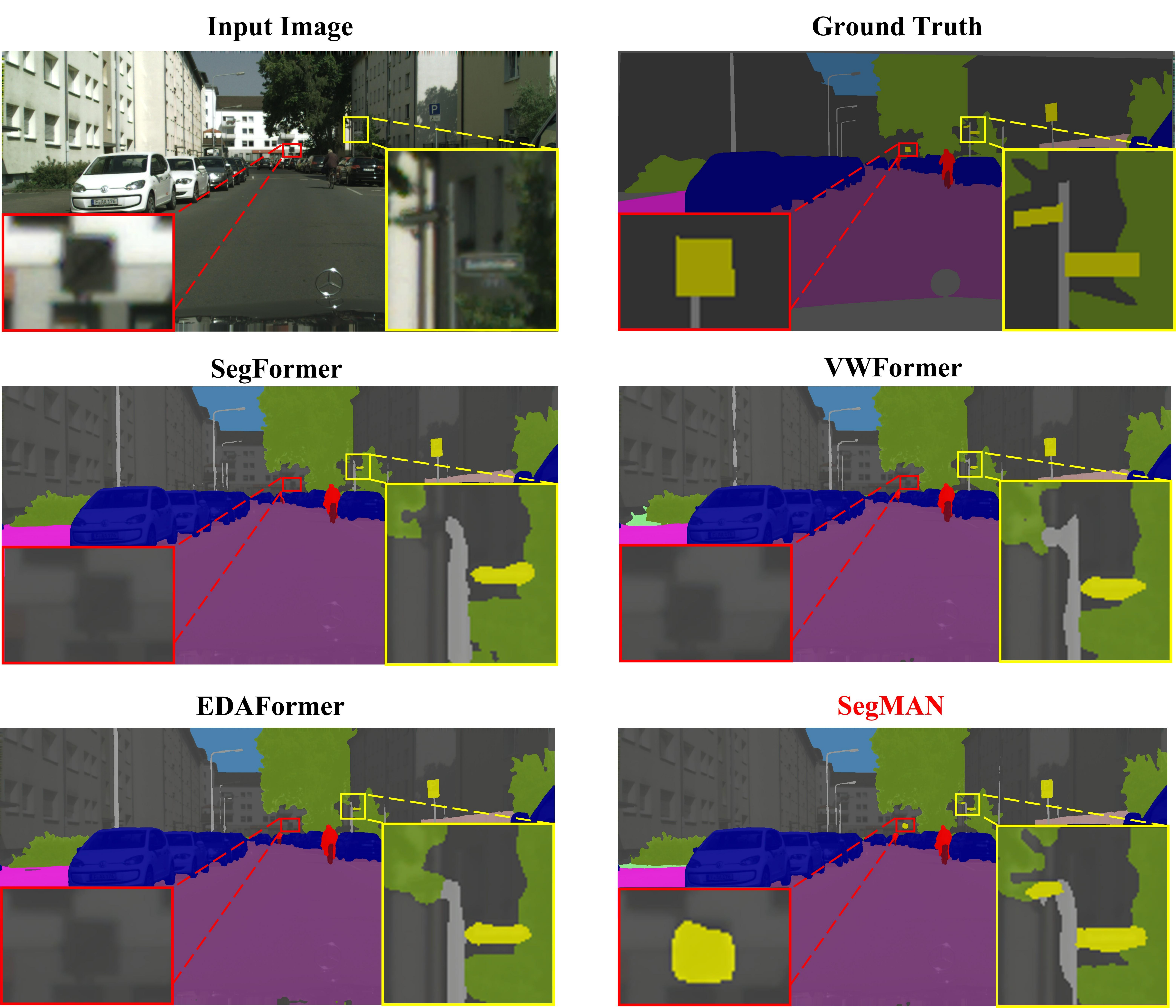}
\label{intro_segmentation_maps}
\end{minipage}
\vspace{-2.5mm}
\caption{Qualitative analysis of receptive field patterns and segmentation performance for small-sized models (27M-29M parameters). \textit{Left}: Visualization of effective receptive fields (ERF) on the Cityscapes validation set at 1024×1024 resolution, illustrating SegMAN's stronger global context modeling capacity in comparison to existing state-of-the-art models. \textit{Right}: Segmentation maps highlighting SegMAN's superior capacity to encode fine-grained local details that are often missed by existing approaches.\vspace{-3mm}}
\label{intro_visualization}
\end{figure*}

As a core computer vision task, semantic segmentation needs to assign a categorical label to every pixel within an image \cite{minaee2021image}. 
Accurate semantic segmentation demands three crucial capabilities. First, \textbf{global context modeling} establishes rich contextual dependencies regardless of spatial distance, enabling overall scene understanding \cite{fu2019dual}. 
Second, \textbf{local detail encoding} endows fine-grained feature and boundary representations, crucial for differentiating semantic categories and localizing boundaries between adjacent regions with different semantic meanings \cite{yu2018learning}.
Third, \textbf{context modeling based on multiple intermediate scales} promotes semantic representations across multiple scales, addressing intra-class scale variations while enhancing inter-class discrimination~\cite{zhao2017pyramid,chen2017deeplab,yan2024vwaformer}.
\par
Recent methods of semantic segmentation have been unable to simultaneously encapsulate all three of these capabilities effectively. For instance, VWFormer~\cite{yan2024vwaformer} introduces a varying window attention (VWA) mechanism that effectively captures multi-scale information through cross-attention between local windows and multiple windows with predefined scales. 
However, at higher input resolutions, the global context modeling capability of VWA diminishes because the predefined window sizes fail to maintain full feature map coverage. In addition, larger windows significantly increase the computational cost as self-attention exhibits quadratic complexity. Likewise, EDAFormer~\cite{yu2024edaformer} proposes an embedding-free spatial reduction attention (SRA) mechanism to implement global attention efficiently and an all-attention decoder for global and multi-scale context modeling. 
However, fine-grained details are lost due to its reliance on downsampled features for token-to-region attention.
Moreover, the absence of dedicated local operators in the feature encoder limits fine-grained feature learning~\cite{xie2021segformer,yu2024edaformer}.
\par

To present a more intuitive understanding of the aforementioned issues, we visualize the effective receptive field (ERF) maps and segmentation maps of recent state-of-the-art models~\cite{xie2021segformer,yan2024vwaformer,yu2024edaformer} with 27M-29M parameters. 
As depicted in Figure~\ref{intro_visualization} (\textit{Left}), the ERF of VWFormer~\cite{yan2024vwaformer} and EDAFormer~\cite{yu2024edaformer} has limited feature map coverage when processing high-resolution Cityscapes images. 
Meanwhile, the segmentation maps in Figure~\ref{intro_visualization} (\textit{Right}) reveal that existing methods have a limited ability to recognize fine-grained local details, which we attribute to insufficient local feature modeling in these architectures.
\par
In this work, we aim to encapsulate omni-scale context modeling for a varying input resolution within a semantic segmentation network. As a result, we propose SegMAN, a novel segmentation network capable of carrying out efficient global context modeling, high-quality local detail encoding, and rich context representation at diverse scales simultaneously. Our SegMAN introduces two novel sub-networks: (1) a hybrid feature encoder that integrates Local Attention and State Space models (LASS) in the token mixer, and (2) a decoder with a Mamba-based Multi-Scale Context Extraction (MMSCopE) module.
In the SegMAN Encoder, LASS leverages a two-dimensional state space model (SS2D)~\cite{liu2024vmamba} for global context modeling and neighborhood attention~\cite{hassani2023neighborhoodAttn} (Natten) for local detail encoding, both functioning in linear time.
The dynamic global scanning in our method always covers the full feature map, enabling robust global context modeling across varying input resolutions and all encoder layers. Meanwhile, by performing local attention within a sliding window, Natten exhibits an outstanding capability in fine-grained local detail encoding.
Our decoder complements the encoder with robust feature learning across multiple scales. The core MMSCopE module first aggregates semantic context from multiple regions of the encoder feature map. Then it uses SS2D to scan each regionally aggregated feature map to learn multi-scale contexts at linear complexity.
For higher efficiency, the feature maps at different scales are reshaped to the same resolution and concatenated so that scanning only needs to be performed on a single feature map. 
Note that our method differs from existing methods in two aspects. First, unlike multi-kernel convolution~\cite{chen2017deeplab} and multi-scale window attention~\cite{yan2024vwaformer}, our design adaptively scales with the input resolution.
Second, by performing SS2D-based scanning over entire multi-scale feature maps, our method better preserves fine-grained details that are typically compromised by pooling operations~\cite{ni2024cgrseg,zhao2017pyramid} or spatial reduction attention~\cite{yu2024edaformer} in existing methods.

As shown in Figure~\ref{segman_performance}, our SegMAN Encoder shows superior performance compared to representative vision backbones on ImageNet-1k. When paired with our novel decoder, the full SegMAN model establishes new state-of-the-art segmentation performance on ADE20K~\cite{zhou2017ade20k}, Cityscapes~\cite{cordts2016cityscapes}, and COCO-Stuff-164k~\cite{caesar2018cocostuff}. 
To summarize, our contributions are threefold:
\begin{itemize}
    \item We introduce a novel encoder architecture featuring our LASS token mixer that synergistically combines local attention with state space models for efficient global context modeling and local detail encoding.
    \item We propose MMSCopE, a novel decoder module that operates on multi-scale feature maps that adaptively scale with the input resolution, surpassing previous approaches in both fine-grained detail preservation and omni-scale context learning.
    \item We demonstrate through comprehensive experiments that SegMAN, powered by LASS and MMSCopE, establishes new state-of-the-art performance while maintaining competitive computational efficiency across multiple challenging semantic segmentation benchmarks.
\end{itemize}

\section{Related work}

\textbf{Segmentation encoders.}
Prior research in semantic segmentation architectures typically comprises two key components: an encoder for feature extraction from input images and a decoder for multi-scale context learning and pixel-wise classification. While conventional approaches often adapt popular classification networks such as ResNet~\cite{he2016deep}, ConvNeXt~\cite{liu2022convnext}, PVT~\cite{wang2021pyramid}, and Swin-Transformer~\cite{liu2021swin}, along with advanced backbones~\cite{shi2024transnext,lou2023transxnet,lou2024sparx,lou2025overlock} as encoders, improvements in classification performance do not necessarily translate to enhanced segmentation capabilities~\cite{he2019bag}, primarily due to the requirement for fine-grained spatial information in segmentation tasks.
This observation has motivated the development of specialized segmentation-oriented encoders~\cite{zheng2021SETR,ranftl2021vision,xie2021segformer,guo2022segnext,yang2022lite,tang2023dynamic,yu2024edaformer,xia2024vit}. 
In this paper, we introduce a novel segmentation encoder that integrates local attention with dynamic State Space Models (Mamba)~\cite{gu2023mamba}. In contrast to previous works, our design enables the simultaneous modeling of fine-grained details and global contexts while maintaining linear time complexity.

\textbf{Segmentation decoders.}
Multi-scale feature learning plays a fundamental role in semantic segmentation, leading to numerous innovations in decoder architectures. Seminal works established various approaches to multi-scale feature aggregation, including the adaptive pooling strategy of PSPNet~\cite{zhao2017pyramid}, atrous convolutions in the DeepLab series~\cite{chen2017deeplab,Chen2017RethinkingAC, chen2018encoder}, and feature pyramid network (FPN) based decoders such as UPerNet~\cite{xiao2018unified} and Semantic FPN~\cite{kirillov2019panoptic}.
Contemporary architectures have advanced both efficiency and performance through diverse approaches to multi-scale feature fusion, including MLP-based methods (Segformer~\cite{xie2021segformer}), Matrix decomposition modules (SegNeXt~\cite{geng2021hamburger, guo2022segnext}), Transformers (FeedFormer~\cite{shim2023feedformer}, VWFormer~\cite{yan2024vwaformer}), and convolutions (CGRSeg~\cite{ni2024cgrseg}). 
An approach relevant to ours is VWFormer~\cite{yan2024vwaformer}, which employs cross-attention between local windows and multiple larger windows with predefined scales to capture multi-scale contexts. However, its reliance on fixed window scales limits its global context modeling capability: as the input resolution increases, the predefined windows fail to maintain complete feature map coverage. 
In contrast, our proposed decoder adaptively extracts context features at different scales while maintaining full feature map coverage regardless of the input resolution.

\textbf{Vision Mamba.} Dynamic state space models represented by Mamba~\cite{gu2023mamba} have demonstrated promising performance in sequence modeling. Mamba combines structured state space models~\cite{gu2021combining,smith2022simplified} with selective scanning, enabling global context modeling with dynamic weights and linear time complexity.
These favorable properties have led to promising results on vision tasks. For example, VMamba~\cite{liu2024vmamba} and ViM~\cite{zhu2024vim} are Mamba-based vision backbone networks. 
Other works have explored Mamba for medical imaging~\cite{liu2024swin,ruan2024vm}, 3D point clouds~\cite{zhang2024point,liu2024point}, and image/video generation~\cite{fei2024DiS,teng2024dim,fu2024lamamba}.
In this paper, we leverage Mamba to learn global contexts in the encoder and multi-scale contexts in the decoder for segmentation.

\section{Method}

\begin{figure*}[t]
\centering
\includegraphics[width=0.92\textwidth]{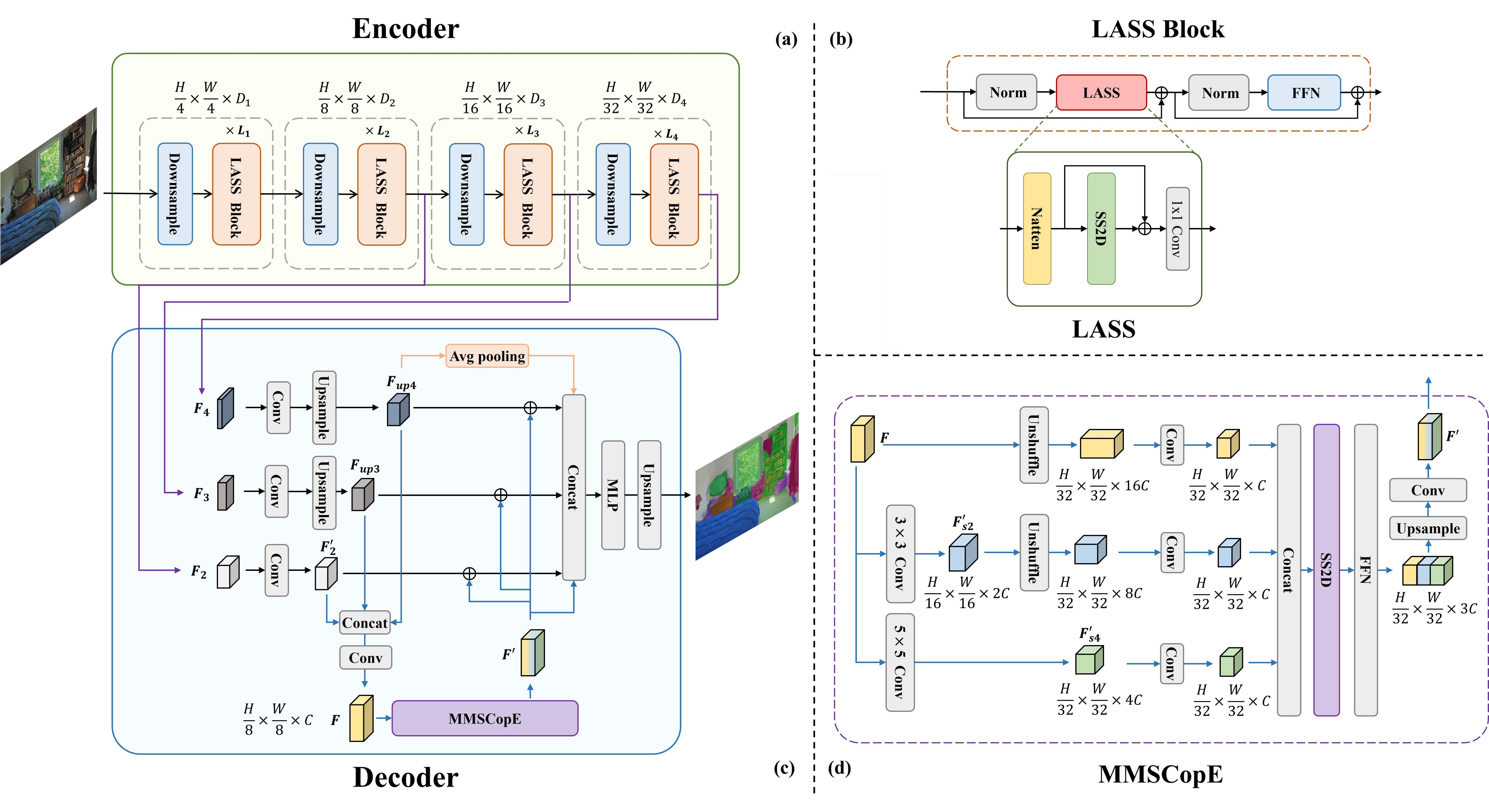}
\caption{Overall Architecture of SegMAN. (a) Hierarchical SegMAN Encoder. (b) LASS for modeling global contexts and local details with linear complexity. (c) The SegMAN Decoder. (d) The MMSCopE module for multi-scale contexts extraction.}
\vspace{-2mm}
\label{method_model}
\end{figure*}

\subsection{Overall Architecture}
As shown in Figure~\ref{method_model}, our proposed SegMAN consists of a newly designed feature encoder and decoder. Specifically, we propose a novel hybrid feature encoder based on core mechanisms from both Transformer and dynamic State Space Models~\cite{gu2023mamba}, and a Mamba-based Multi-Scale Context Extraction module (MMSCopE) in the decoder. The hierarchical SegMAN Encoder integrates Neighborhood Attention (Natten)~\cite{hassani2023neighborhoodAttn} and the 2D-Selective-Scan Block (SS2D in VMamba)~\cite{liu2024vmamba} within the token mixer termed Local Attention and State Space (LASS), enabling comprehensive feature learning at both global and local scales across all layers while maintaining linear computational complexity. Meanwhile, MMSCoPE dynamically and adaptively extracts multi-scale semantic information by processing feature maps at varying levels of granularity using SS2D~\cite{liu2024vmamba}, with the scales adaptively adjusted according to the input resolution. Collectively, the SegMAN Encoder injects robust global contexts and local details into a feature pyramid, where feature maps are progressively downsampled and transformed to produce omni-scale features at different pyramid levels. This feature pyramid is then fed into MMSCopE, where features are further aggregated spatially as well as across scales, resulting in comprehensive multi-scale representations that can be used for dense prediction.

\subsection{Feature encoder}
\textbf{Overview.}
As shown in Figure~\ref{method_model}, our SegMAN Encoder is a standard four-stage network~\cite{liu2021swin,liu2022convnext}, where each stage begins with a strided 3$\times$3 convolution for spatial reduction, followed by a series of LASS Blocks. Each LASS Block includes pre-layer normalization, a novel LASS module, and a feedforward network (FFN)~\cite{dosovitskiy2020vit, liu2021swin,hassani2023neighborhoodAttn}.
Our proposed LASS represents the first integration of local self-attention (Natten~\cite{hassani2023neighborhoodAttn}) and state space models (SS2D~\cite{liu2024vmamba}) for semantic segmentation, capturing local details and global contexts simultaneously with linear time complexity. Ablation studies (Table~\ref{tab:ablation_backbone}) confirm the necessity of both components.

\textbf{Local attention and state space module.}
Long-range dependency modeling is crucial for semantic segmentation as they enable comprehensive context learning and scene understanding, as demonstrated by the success of Transformer-based semantic segmentation models~\cite{zheng2021SETR,ranftl2021vision,xie2021segformer,lu2023content,yu2024edaformer}.
To efficiently learn global contexts across all network layers, we leverage the SS2D block in VMamba~\cite{liu2024vmamba}. This variant of a dynamic state space model named Mamba~\cite{gu2023mamba} adopts four distinct and complementary scanning paths, enabling each token to integrate information from all other tokens in four different directions.

However, while Mamba~\cite{gu2023mamba} achieves linear time complexity by compressing the global context of each channel into a fixed-dimensional hidden state~\cite{gu2023mamba}, this compression inherently leads to information loss, in particular, loss of fine-grained local spatial details, that are crucial in segmentation to accurately localize region boundaries. To this end, we utilize Neighborhood Attention (Natten)~\cite{hassani2023neighborhoodAttn}, where a sliding-window attention mechanism localizes every pixel's attention span to its immediate neighborhood. This approach retains translational equivalence, effectively captures local dependencies, and has linear time complexity. In practice, we serially stack Natten and SS2D, and a shortcut around SS2D is employed to merge local and global information. The merged output is then fed into a 1$\times$1 convolution layer for further global-local feature fusion.

\begin{table}[t]
\centering
\scalebox{0.75}{
\begin{tabular}{ccccc}
\toprule
SegMAN & \multirow{2}{*}{Channels} & \multirow{2}{*}{Blocks} & Params & FLOPs \\
Encoder & & & (M) & (G) \\
\midrule
\midrule
Tiny & [32, 64, 144, 192] & [2, 2, 4, 2] & 3.5 & 0.65 \\
Small & [64, 144, 288, 512] & [2, 2, 10, 4] & 25.5 & 4.05 \\
Base & [80, 160, 364, 560] & [4, 4, 18, 4] & 45.0 & 9.94 \\
Large & [96, 192, 432, 640] & [4, 4, 28, 4] & 81.0  & 16.8 \\
\bottomrule
\end{tabular}}
\caption{Configurations of the three SegMAN Encoder variants. FLOPs were measured at the \(224 \times 224\) resolution.\vspace{-3mm}}
\label{method_model_config}
\end{table}

\textbf{Network architecture.}
Our SegMAN Encoder produces a feature pyramid. The spatial resolution of \(F_i\) is \(\frac{H}{2^{1+i}} \times \frac{W}{2^{1+i}} \), where \(H\) and \(W\) are the height and width of the input image, respectively. We design three SegMAN Encoder variants with different model sizes, each with computational complexity comparable to some existing semantic segmentation encoders~\cite{xie2021segformer,guo2022segnext,yu2024edaformer}. 
As the feature map resolution at Stage 4 is (\(\frac{H}{32} \times \frac{W}{32}\)), global self-attention~\cite{vaswani2017attention} becomes computationally feasible. Therefore we replace SS2D with this more powerful global context modeling mechanism.
The configurations of these models are presented in Table~\ref{method_model_config}.

\subsection{Feature Decoder}
\textbf{Overview.} As shown in Figure~\ref{method_model}, our proposed decoder first aggregates features at various levels of abstraction (from low-level \(F_2\) to high-level \(F_4\)) to obtain a feature map \(F\) of size \(\frac{H}{8} \times \frac{W}{8} \), as in prior works~\cite{xie2021segformer,yan2024vwaformer}.
Specifically, each feature map \(F_i\) is projected to a lower dimension using a \(1\times 1\) convolution followed by batch normalization and ReLU, collectively denoted as `Conv' in Figure~\ref{method_model}. 
Subsequently, \(F_3\) and \(F_4\) are upsampled using bilinear interpolation to match the spatial dimensions of \(F_2\). 
The upsampled features \(F_{up3}\) and \(F_{up4}\) are concatenated together with \(F_2\) and passed through a Conv layer for channel dimension reduction, resulting in \(F\) with size \(\frac{H}{8} \times \frac{W}{8} \times C\), where \(C\) is the number of channels. 
The feature map \(F\) is then processed by our proposed MMSCopE module to extract rich multi-scale context features, producing a new feature map \(F'\).
We then add the feature map \(F'\) to \(F_{2}\), \(F_{up3}\), and \(F_{up4}\) to enrich their multi-scale contexts. The final prediction pathway concatenates \(F'\) with these context-enhanced features and the global pooling result on \(F_{up4}\), passes the combined features through a two-layer MLP, and the result is upsampled to the input resolution.

\textbf{Mamba-based multi-scale context extraction.}
To effectively extract contexts at different scales, we propose a Mamba-based Multi-Scale Context Extraction (MMSCopE) module. As shown in Figure~\ref{method_model}, we apply strided convolutions to downsample the feature map \(F\), generating \(F_{s2}\) and \(F_{s4}\) with a resolution equal to 1/2 and 1/4 of that of \(F\), respectively.
The motivation behind this is to obtain multiple regionally aggregated contexts through two derived feature maps: \(F_{s2}\), which aggregates features from \(3\times 3\) neighborhoods using a convolution with stride 2, and \(F_{s4}\), which aggregates features from \(5\times 5\) neighborhoods with stride 4.
Then, motivated by the observation that Mamba scans each channel independently, we use a single Mamba scan to simultaneously extract contexts from the three feature maps \(F\), \(F_{s2}\), and \(F_{s4}\).
The main idea is to concatenate these feature maps along the channel dimension and perform a single Mamba scan, which is more efficient on the GPU than processing each feature map separately.
However, the spatial dimensions of these feature maps are not consistent, which prevents direct concatenation. To resolve this, we employ lossless downsampling via the \emph{Pixel Unshuffle}~\cite{shi2016real} operation to reduce the spatial dimensions of \(F\) and \(F_{s2}\) to match that of \(F_{s4}\), which is \(\frac{H}{32} \times \frac{W}{32}\). 
Specifically, this operation rearranges non-overlapping 4×4 patches from \(F\) and 2×2 patches from \(F_{s2}\) into their respective channel dimension, increasing the channel depth by a factor of 16 and 4 respectively, while preserving complete spatial information. 
The transformed feature maps retain information at their original scales.
To reduce computational costs, we project each feature map to a fixed channel dimension \(C\) through \(1\times 1\) convolutions. This step assigns equal importance to the context information from each scale.
Then we concatenate the projected feature maps along the channel dimension, and pass the combined feature map to SS2D~\cite{liu2024vmamba} to achieve multi-scale context extraction in a single scan.
The resulting feature map is of size \(\frac{H}{8} \times \frac{W}{8} \times 3C\), where the three segments along the channel dimension correspond to earlier feature maps at three distinct scales (\(F\), \(F_{s2}\), and \(F_{s4}\)).
To facilitate context mixing across scales, we append a \(1\times 1\) convolution layer after the SS2D scan.
Finally, we use bilinear interpolation to upsample the mixed features to 1/8 of the input resolution, and project the channel dimension down to \(C\) using another \(1\times 1\) convolution. This yields the new feature map \(F'\) with mixed multi-scale contexts.

\textbf{Multi-scale fusion.}
Instead of directly using the feature map \(F'\) to predict pixel labels, we further exploit stage-specific representations from the encoder by adding \(F'\) to \(F_{2}\), \(F_{up3}\), and \(F_{up4}\). This effectively injects multi-scale contexts into stage-specific feature maps, which inherits information from various levels of abstraction.
Next, we concatenate the resulting context-enhanced feature maps from each stage, the multi-scale context feature \(F'\), and the global average pooling result on \(F_{up4}\).
For pixel-wise label prediction, the diverse concatenated features are fused through a two-layer MLP, and further bilinearly interpolated to restore the original input resolution.

\newcommand{\cmark}{\ding{51}}%
\newcommand{\xmark}{\ding{55}}%

\definecolor{posblue}{RGB}{0, 91, 187}
\definecolor{red}{RGB}{205, 92, 92}

\begin{table*}[t]
\centering

\begin{minipage}[t]{0.6\textwidth} 
\centering
\resizebox{0.99\textwidth}{!}{ 
\begin{tabular}{@{}lc|cc|cc|cc@{}}
\toprule
\multirow{2}{*}{Method} & \multirow{2}{*}{Params(M)} & \multicolumn{2}{c|}{ADE20K} & \multicolumn{2}{c|}{Cityscapes} & \multicolumn{2}{c}{COCO-Stuff} \\
& & GFLOP & mIoU & GFLOP & mIoU & GFLOP & mIoU \\
\midrule
\midrule
\midrule

Segformer-B0~\cite{xie2021segformer} & 3.8 & 8.4 & 37.4 & 125.5 & 76.2 & 8.4 & 35.6 \\
SegNeXt-T~\cite{guo2022segnext} & 4.3 & 7.7 & 41.1 & 61.6 & 78.9 & 7.7 & 38.7 \\
VWFormer-B0~\cite{yan2024vwaformer} & 3.7 & 5.8 & 38.9 & 112.4 & 77.2 & 5.8 & 36.2 \\
EDAFormer-T~\cite{yu2024edaformer} & 4.9 & 5.8 & 42.3 & 151.7 & 78.7 & 5.8 & 40.3 \\
CGRSeg-T$^{\dagger}$~\cite{ni2024cgrseg} & 9.4 & 4.8 & 42.5 & 65.5 & 78.3 & 4.8 & 40.4 \\
\rowcolor{ggray} SegMAN-T & 6.4 & 6.2 & \textbf{43.0} & 52.5 & \textbf{80.3} & 6.2 & \textbf{41.3} \\

\midrule
Swin UperNet-T~\cite{liu2021swin} & 60.0 & 236.0 & 44.4 & - & - & - & - \\
ViT-CoMer-S~\cite{xia2024vit} & 61.4 & 296.1 & 46.5 & - & - & - & - \\
OCRNet~\cite{yuan2020ocrnet} & 70.5 & 164.8 & 45.6 & - & - & - & -\\
Segformer-B2~\cite{xie2021segformer} & 27.5  & 62.4 & 46.5 & 717.1 & 81.0 & 62.4 & 44.6 \\
MaskFormer~\cite{cheng2021maskformer} & 42.0 & 55.0 & 46.7 & - & - & - & - \\
Mask2Former~\cite{cheng2022masked2former} & 47.0 & 74.0 & 47.7 & - & - & - & - \\
SegNeXt-B~\cite{guo2022segnext} & 27.6 & 34.9 & 48.5 & 279.1 & 82.6 & 34.9 & 45.8 \\
FeedFormer-B2~\cite{shim2023feedformer} & 29.1 & 42.7 & 48.0 & 522.7 & 81.5 & -& - \\
VWFormer-B2~\cite{yan2024vwaformer} & 27.4 & 46.6 & 48.1 & 415.1 & 81.7 & 46.6 & 45.2 \\
EDAFormer-B~\cite{yu2024edaformer} & 29.4 & 32.1 & 49.0 & 605.9 & 81.6 & 32.1 & 45.9 \\
CGRSeg-T$^{\dagger}$~\cite{ni2024cgrseg} & 35.7 & 16.5 & 47.3 & 199.7 & 80.2 & 16.5 & 45.2 \\
\rowcolor{ggray} SegMAN-S & 29.4 & 25.3 & \textbf{51.3} & 218.4 & \textbf{83.2} & 25.3 & \textbf{47.5}\\

\midrule
Segformer-B3~\cite{xie2021segformer} & 47.3 & 79.0 & 49.4 & 962.9 & 81.7 & 79.0 & 45.5 \\
SegNeXt-L~\cite{guo2022segnext} & 48.9 & 70.0 & 51.0 & 577.5 & 83.2 & 70.0 & 46.5 \\
VWFormer-B3~\cite{yan2024vwaformer} & 47.3 & 63.3 & 50.3 & 637.1 & 82.4 & 63.3 & 46.8 \\
\rowcolor{ggray} SegMAN-B & 51.8 & 58.1 & \textbf{52.6} & 479.0 & \textbf{83.8} & 58.1 & \textbf{48.4} \\

\midrule
Swin UperNet-B~\cite{liu2021swin} & 121.0 & 261.0 & 48.1 & - & - & - & - \\
ViT-CoMer-B~\cite{xia2024vit} & 144.7 & 455.4 & 48.8 & - & - & - & - \\
Segformer-B5~\cite{xie2021segformer} & 84.7 & 110.3 & 51.0 & 1149.8 & 82.4 & 110.3 & 46.7 \\
VWFormer-B5~\cite{yan2024vwaformer} & 84.6 & 96.1 & 52.0 & 1139.6 & 82.8 & 96.1 & 48.0 \\
\rowcolor{ggray} SegMAN-L & 92.4 & 97.1 & \textbf{53.2} & 796.0 & \textbf{84.2} & 97.1 & \textbf{48.8} \\
\bottomrule
\end{tabular}}
\caption{Comparison with state-of-the-art semantic segmentation models on ADE20K, Cityscapes, and COCO-Stuff-164K. FLOPs are calculated at \(512\times 512\) (ADE20K, COCO-Stuff) and \(2048\times 1024\) (Cityscapes) resolutions.}
\label{exp_main_segmentation}
\end{minipage}
\hspace{3mm}
\begin{minipage}[t]{0.342\textwidth} 
\centering
\resizebox{0.99\textwidth}{!}{ 
\begin{tabular}{lccc}
\toprule
Models & Params(M) & GFLOP & Acc \\
\midrule
\midrule
\midrule
MiT-B0~\cite{xie2021segformer} & 3.8 &0.60 & 70.5 \\
EFT-T~\cite{yu2024edaformer} &3.7 &0.60 & 72.3 \\ 
MSCAN-T~\cite{guo2022segnext} & 4.2 & 0.89 & 75.9 \\
\rowcolor{ggray} SegMAN-T Encoder & 3.5 & 0.65 & \textbf{76.2} \\
\midrule
MiT-B2~\cite{xie2021segformer} & 24 & 4.0 & 81.6 \\
EFT-B~\cite{yu2024edaformer} & 26 & 4.2 & 82.4 \\
MSCAN-B~\cite{guo2022segnext} & 27 & 4.4 & 83.0 \\
Swin-T~\cite{liu2021swin} & 28 & 4.5 & 81.2\\
ConvNeXt-T~\cite{liu2022convnext} & 28 & 4.5 & 82.1 \\
InternImage-T~\cite{wang2023internimage} & 30 & 5.0 & 83.5 \\
%
ViM-S~\cite{zhu2024vim} & 26 & - & 81.6 \\
VMamba-T~\cite{liu2024vmamba} & 29 & 4.9 & 82.6 \\
SparX-Mamba-T~\cite{lou2024sparx} & 27 & 5.2 & 83.5 \\
\rowcolor{ggray} SegMAN-S Encoder & 26 & 4.1 & \textbf{84.0} \\
\midrule
MiT-B3~\cite{xie2021segformer} & 45 & 6.9 & 83.1 \\
MSCAN-L~\cite{guo2022segnext} & 45 & 9.1 & 83.9 \\
Swin-S~\cite{liu2021swin} & 50 & 8.7 & 83.2\\
ConvNeXt-S~\cite{liu2022convnext} & 50 & 8.7 & 83.1 \\
InternImage-S~\cite{wang2023internimage} & 50 & 8.0 & 84.2 \\
%
VMamba-S~\cite{liu2024vmamba} & 50 & 8.7 & 83.6 \\
SparX-Mamba-S~\cite{lou2024sparx} & 47 & 9.3 & 84.2 \\
\rowcolor{ggray} SegMAN-B Encoder & 45 & 9.9 & \textbf{85.1} \\
\midrule
MiT-B5~\cite{xie2021segformer} & 82 & 11.8 & 83.8 \\
Swin-B~\cite{liu2021swin} & 88 & 15.4 & 83.5\\
ConvNeXt-B~\cite{liu2022convnext} & 89 & 15.4 & 83.8 \\
InternImage-B~\cite{wang2023internimage} & 97 & 16.7 & 84.9 \\
%
ViM-B~\cite{zhu2024vim} & 98 & - & 83.2 \\
VMamba-B~\cite{liu2024vmamba} & 89 & 15.4 & 83.9 \\
SparX-Mamba-B~\cite{lou2024sparx} & 84 & 15.9 & 84.5 \\
\rowcolor{ggray} SegMAN-L Encoder & 81 & 16.8 & \textbf{85.5} \\
\bottomrule
\end{tabular}}
\caption{Comparison with state-of-the-art vision backbones on ImageNet-1K. FLOPs are calculated at \(224\times 224\) resolution.}
\label{exp_encoder}
\end{minipage}

\vspace{-3mm} 
\end{table*}

\section{Experiments}

\subsection{Setting}
\textbf{Datasets.}
Following previous works \cite{xie2021segformer,guo2022segnext}, we firstly pre-train our SegMAN Encoder on the ImageNet-1K dataset~\cite{deng2009imagenet}. Then, we evaluate the semantic segmentation performance on three commonly used datasets: ADE20K~\cite{zhou2017ade20k}, a challenging scene parsing benchmark consisting of 20,210 images annotated with 150 semantic categories; Cityscapes~\cite{cordts2016cityscapes}, an urban driving dataset comprising 5,000 high-resolution images with 19 semantic categories; and COCO-Stuff-164K~\cite{caesar2018cocostuff}, which contains 164,000 images labeled across 172 semantic categories.

\textbf{Implementation details.}
We conducted ImageNet pre-training using the Timm library~\cite{rw2019timm} with hyperparameters and data augmentation settings identical to those of Swin Transformer~\cite{liu2021swin}. Regarding semantic segmentation, we trained all models using the MMSegmentation library~\cite{mmseg2020}, combining the pre-trained SegMAN Encoder with a randomly initialized decoder.
Following the Segformer training protocol~\cite{xie2021segformer}, we employed standard data augmentation techniques, including random horizontal flipping and random scaling (ratio of 0.5 to 2.0). We implemented dataset-specific random cropping with dimensions of 512×512 for ADE20K and COCO-Stuff, and 1024×1024 for Cityscapes. We trained the models using the AdamW optimizer for 160K iterations with batch sizes of 16 for ADE20K and COCO-Stuff, and 8 for Cityscapes. The learning rate is set to 1e-6 with 1500 warmup iterations.
We evaluated segmentation performance using mean intersection over union (mIoU) and computed FLOPs using the fvcore library~\cite{fvcore}.
All models were trained on 8 NVIDIA H800 GPUs.

\subsection{Comparison with state-of-the-art methods}
This section presents a comprehensive comparison of our proposed SegMAN with state-of-the-art segmentation methods on three widely used datasets.

\textbf{Baselines.}
We compare our method with other strong competitors including EDAFormer~\cite{yu2024edaformer}, VWFormer~\cite{yan2024vwaformer} and CGRSeg~\cite{ni2024cgrseg}, Segformer~\cite{xie2021segformer}, and SegNeXt~\cite{guo2022segnext}.
Note that the Efficientformer-v2~\cite{li2022efficientformerv2} used in CGRSeg~\cite{ni2024cgrseg} significantly benefits from knowledge distillation from RegNetY-16GF~\cite{radosavovic2020designing} (83.6M parameters, 15.9 GFLOPs) and extended training duration of 450 epochs compared to the conventional 300 epochs, providing a stronger encoder network compared to other methods. To ensure a fair comparison, we retrained Efficientformer-v2 on ImageNet-1K using the same training configurations as Swin~\cite{liu2021swin}. We report results for CGRSeg by fine-tuning this retrained encoder following official implementations.


\textbf{Results.}
Table~\ref{exp_main_segmentation} shows the superior performance of SegMAN across all three datasets: ADE20K, Cityscapes, and COCO-Stuff-164K. Among lightweight models, SegMAN-T achieves the best performance, with 43.0\% mIoU on ADE20K, 80.3\% mIoU on Cityscapes, and 41.3\% mIoU on COCO-Stuff, while maintaining comparable computational costs. 
Our SegMAN-S model achieves significant performance improvements compared to state-of-the-art methods.
On the ADE20K benchmark, it surpasses the recently introduced EDAFormer-B~\cite{yu2024edaformer} by 2.3\% mIoU, and even outperforms the larger SegNeXt-L~\cite{guo2022segnext} model, which has 48.9M parameters, by 0.3\% mIoU. The performance gains extend to the COCO-Stuff dataset, where SegMAN-S demonstrates improvements of 1.6\% mIoU over EDAFormer-B and 1.0\% mIoU over SegNeXt-L. Notably, SegMAN-S achieves these superior results while reducing computational complexity by over 20\%, requiring only 25.3 GFLOPs compared to EDAFormer-B.
On Cityscapes, SegMAN-S achieves 83.2\% mIoU, surpassing EDAFormer-B by 1.6\%.
Moreover, SegMAN-B surpasses previous best performing models by 1.6\% on both ADE20K and COCO-Stuff while incurring lower computational cost. On Cityscapes, SegMAN-B improves SOTA by 0.6\% mIoU using 17\% less GFLOPs.
Similarly, our largest SegMAN-L model achieves +1.2\%, +1.0\%, and +0.8\% gains over prior SOTA on ADE20K, Cityscapes, and COCO-Stuff, respectively.
The consistent performance gains across different datasets and model scales validate the effectiveness of our proposed SegMAN, which can simultaneously capture global contexts, local details, and multi-scale information through its LASS-based encoder and MMSCopE-based decoder. 

\textbf{Speed Analysis.}
We benchmark inference speed on Cityscapes~\cite{cordts2016cityscapes} using an NVIDIA L40S GPU, averaging Frames Per Second (FPS) over 128 steps with batch size 2.
As shown in Table~\ref{tab:latency_comparison}, SegMAN-T achieves +1.6\% mIoU at approximately 3 times the speed of EDAFormer-T, highlighting an excellent balance between performance and speed.

\begin{table}
\setlength{\belowcaptionskip}{-0.2cm}
\centering
\scalebox{0.8}{
\begin{tabular}{lcccc}
\toprule
Model  & Params (M) & GFLOPs & FPS & mIoU \\
\midrule
\midrule
Segformer-B0~\cite{xie2021segformer} & 3.8 & 125.5 & 21.8& 76.2 \\
SegNext-T~\cite{guo2022segnext} & 4.3 & 61.6 & 25.8 & 78.9\\
VWFormer-B0~\cite{yan2024vwaformer} & 3.7 & 112.4 & 21.1 & 77.2\\
EDAFormer-T~\cite{yu2024edaformer} & 4.9 & 151.7 & 12.7 & 78.7 \\
CGRSeg-T~\cite{ni2024cgrseg}  & 9.3 & 65.5 & 27.6 & 78.3 \\
\rowcolor{ggray} SegMAN-T & 6.4 & 52.5 & 34.9 & \textbf{80.3} \\
\midrule
Segformer-B2~\cite{xie2021segformer} & 27.5 & 717.1 & 10.2 & 81.0 \\
SegNext-B~\cite{guo2022segnext} & 27.6 & 279.1 & 14.1 & 82.6 \\
VWFormer-B2~\cite{yan2024vwaformer} & 27.4 & 415.1 & 9.87 & 81.7 \\
EDAFormer-B~\cite{yu2024edaformer} & 29.4 & 605.9 & 6.6 & 81.6 \\
CGRSeg-B~\cite{ni2024cgrseg} & 35.6 & 199.0 & 11.8 & 80.2 \\
\rowcolor{ggray} SegMAN-S & 29.4 & 218.4 & 13.8 & \textbf{83.2} \\

\midrule
Segformer-B3~\cite{xie2021segformer} & 46.3 & 962.9 & 6.7 & 81.7 \\
SegNext-L~\cite{guo2022segnext} & 48.9 & 577.5 & 10.1 & 83.2 \\
VWFormer-B3~\cite{yan2024vwaformer} & 47.3 & 637.1 & 6.3 & 82.4 \\
\rowcolor{ggray} SegMAN-B & 51.8 & 479.0 & 7.2 & \textbf{83.8} \\
\midrule
Segformer-B5~\cite{xie2021segformer} & 84.7 & 1460.4 & 3.5 & 81.7 \\
VWFormer-B5~\cite{yan2024vwaformer} & 84.6 & 1139.6 & 3.7 & 82.4 \\
\rowcolor{ggray} SegMAN-L & 92.4 & 796.0 & 5.2 & \textbf{84.2} \\
\bottomrule
\end{tabular}}
\caption{Comparison of model complexity and inference speed on Cityscapes~\cite{cordts2016cityscapes} (\(2048\times 1024\) resolution).}
\label{tab:latency_comparison}
\end{table}

\subsection{Comparison with state-of-the-art backbones}
In Table~\ref{exp_encoder}, we evaluate the classification performance of our proposed SegMAN Encoder on ImageNet-1K~\cite{deng2009imagenet}, comparing against two categories of architectures: state-of-the-art segmentation encoders (MiT~\cite{xie2021segformer}, MSCAN~\cite{xie2021segformer}, EFT~\cite{yu2024edaformer}) and representative vision backbones (Swin Transformer~\cite{liu2021swin}, ConvNeXt~\cite{liu2022convnext}, InternImage~\cite{wang2023internimage} ViM~\cite{zhu2024vim}, VMamba~\cite{liu2024vmamba}, SparX-Mamba~\cite{lou2024sparx}).
SegMAN Encoder variants consistently achieve superior classification accuracy in comparison to architectures of similar computational complexity. Specifically, SegMAN-S Encoder significantly surpasses ConvNeXt-T and MSCAN-B by 1.9\% and 1.0\% in Top-1 accuracy, respectively. SegMAN-B Encoder demonstrates similar improvements over Swin-S and MSCAN-L by 1.9\% and 1.2\%, respectively.
The promising performance across model variants demonstrates the benefits of integrating local attention with state space models to effectively capture both fine-grained details and global contextual information. A more detailed comparison can be found in the supplementary material.

\begin{table}[t]
\centering
\scalebox{0.66}{
\begin{tabular}{lccccc}
\toprule
\multirow{2}{*}{Model Variant} & Params & \multirow{2}{*}{GFLOPs}  & \multirow{2}{*}{FPS} & \multirow{2}{*}{Acc} & \multirow{2}{*}{mIoU} \\
 & (M) & & & \\
\midrule
\midrule
SegMAN-S Encoder & 25.5 & 21.4 & 139 & 84.0 & 51.3 \\
\midrule
\multicolumn{5}{l}{\textit{Replace LASS}} \\
MaxViT~\cite{tu2022maxvit} & 24.6 & 29.8 & 96  & 83.5 \textcolor{red}{(-0.5)} &  47.2 \textcolor{red}{(-4.1)}\\
ACMix~\cite{pan2022acmix} & 24.9 & 19.3 & 104 & 83.1 \textcolor{red}{(-0.9)} & 48.6 \textcolor{red}{(-2.7)} \\
PVT~\cite{wang2021pyramid} & 29.5 & 22.0 & 169 & 82.8 \textcolor{red}{(-1.2)} & 49.1 \textcolor{red}{(-2.2)} \\
BiFormer~\cite{zhu2023biformer} & 25.2 & 30.5 & 97 & 82.9 \textcolor{red}{(-1.1)} &  48.8 \textcolor{red}{(-2.5)} \\

\midrule
\multicolumn{5}{l}{\textit{LASS Design}} \\
Parallel Natten and SS2D & 25.5 & 21.4 & 129  & 83.8 \textcolor{red}{(-0.2)} & 48.9  \textcolor{red}{(-2.4)} \\
w/o SS2D residual & 26.1 & 21.4 & 132 & 83.7 \textcolor{red}{(-0.3)} & 50.2 \textcolor{red}{(-1.1)} \\
Stage 4 Attn → SS2D & 30.0 & 21.8 & 121 & 83.7 \textcolor{red}{(-0.3)} & 50.6 \textcolor{red}{(-0.7)} \\

\midrule
\multicolumn{5}{l}{\textit{SS2D within LASS}} \\
Replace SS2D w SRA~\cite{wang2021pyramid} & 26.0 & 22.3 & 142 & 83.7 \textcolor{red}{(-0.3)} & 50.6 \textcolor{red}{(-0.7)} \\
Replace SS2D w L-Attn~\cite{katharopoulos2020transformers} & 27.6 & 21.8 & 150 & 83.6 \textcolor{red}{(-0.4)} & 49.5 \textcolor{red}{(-1.8)} \\
Remove SS2D & 25.0 & 20.9 & 162  & 83.6 \textcolor{red}{(-0.4)} & 47.4 \textcolor{red}{(-3.9)} \\
\midrule

\multicolumn{5}{l}{\textit{Natten within LASS}} \\
Replace Natten w Conv & 25.8 & 20.7 & 160 & 83.1 \textcolor{red}{(-0.9)} & 49.5 \textcolor{red}{(-1.8)}\\
Replace Natten w Swin~\cite{liu2021swin} & 30.0 & 21.6 & 147 & 83.5 \textcolor{red}{(-0.5)} & 50.3 \textcolor{red}{(-1.0)} \\
Remove Natten~\cite{hassani2023neighborhoodAttn} & 24.9 & 25.9 & 100 & 83.6 \textcolor{red}{(-0.4)} & 49.8 \textcolor{red}{(-1.5)} \\

\bottomrule
\end{tabular}}
\caption{Ablation studies and comparison of token mixers for SegMAN encoder. FLOPs and FPS are calculated or measured at \(512\times 512\) resolution using the encoder.\vspace{-3mm}}
\label{tab:ablation_backbone}
\end{table}

\subsection{Ablation studies on SegMAN encoder}
We systematically evaluate SegMAN Encoder's architectural design on ImageNet-1K~\cite{deng2009imagenet} and ADE20K~\cite{zhou2017ade20k}. Our LASS token mixer integrates Neighborhood Attention (Natten~\cite{hassani2023neighborhoodAttn}) and 2D-Selective-Scan (SS2D), with a residual connection added to SS2D to fuse global and local contexts. In Stage 4 blocks, we replace SS2D with global attention for enhanced high-level feature modeling.

To validate these architectural decisions, we conduct comprehensive ablation studies examining: (1) comparative performance against alternative token mixers, (2) design choices of LASS, and (3) the impact of individual components within the token mixer. In all experiments, including token mixer replacements, we maintain consistent model complexity by adjusting channel dimensions to achieve comparable parameter counts and computational costs. Table~\ref{tab:ablation_backbone} presents our findings.

\textbf{Alternative token mixers.}
We evaluate the effectiveness of LASS by replacing it with recent token mixers (MaxViT~\cite{tu2022maxvit}, ACMix~\cite{pan2022acmix}, PVT~\cite{wang2021pyramid}, and BiFormer~\cite{zhu2023biformer}). Note that we implement MaxViT's Block-SA and Grid-SA modules in series to serve as a token mixer. As shown in Table~\ref{tab:ablation_backbone}, our LASS achieves consistent and substantial improvements across both classification and segmentation tasks while maintaining competitive computational costs. The performance gains stem from the complementary nature between Natten, which models fine-grained local features, and SS2D, which efficiently models global contexts.

\textbf{Alternative LASS mixer designs.} We explore alternative methods to integrate SS2D and Natten together in LASS. Results indicate that parallel arrangement of Natten and SS2D and removing the residual connections for SS2D lead to substantial performance drops in both classification accuracy and segmentation mIoU.

\textbf{SS2D and Natten.}
We analyze the two key components of LASS through systematic ablations. When SS2D is replaced with alternative global attention mechanisms, performance consistently decreases: SRA~\cite{wang2021pyramid} (-0.3\% accuracy, -0.7\% mIoU) and Linear Attention~\cite{katharopoulos2020transformers} (-0.4\% accuracy, -1.8\% mIoU). Likewise, replacing Natten with either convolutions (-0.9\% accuracy, -1.8\% mIoU) or Shifted-Window Attention~\cite{liu2021swin} (-0.5\% accuracy, -1.0\% mIoU) leads to significant performance degradation, demonstrating the effectiveness of our component choices.

\subsection{Ablation studies on SegMAN decoder}
\label{main_ablation_decoder_section}
We investigate both the overall effectiveness of our proposed decoder and the performance of individual components of the MMSCopE module using the ADE20K dataset~\cite{zhou2017ade20k}.
For a fair comparison, we ensure that different model variants have comparable computational complexity by adjusting the number of channels. Here, all ablations are conducted using the SegMAN-S decoder.


\begin{table}[t]
\centering
\scalebox{0.77}{
\begin{tabular}{l c c c}
\toprule
Model & Params (M) & GFLOPs  & mIoU (\%) \\
\midrule
UPerNet~\cite{xiao2018unified} & 27.1 & 38.7 & 50.4 \\
MLP Decoder~\cite{xie2021segformer} & 25.0 & 32.7 & 50.3 \\
Ham Decoder~\cite{guo2022segnext} & 25.3 & 29.5 & 50.4 \\
VWFormer Decoder~\cite{yan2024vwaformer} & 28.2 & 40.7 & 50.5 \\
CGRHead~\cite{ni2024cgrseg} & 44.2 & 24.5 & 50.3 \\
EDAFormer Decoder~\cite{yu2024edaformer} & 28.5 & 26.2 & 50.2 \\
\rowcolor{ggray}SegMAN-S Decoder & 29.4 & \textbf{25.3}  & \textbf{51.3} \\
\bottomrule
\end{tabular}}
\caption{Comparison of complexity and accuracy of segmentation decoders. FLOPs are calculated at \(512\times 512\) resolution.\vspace{-6mm}} 
\label{tab:ablation_head_replace}
\end{table}

\textbf{Decoder comparisons and encoder generalization}.
Table~\ref{tab:ablation_head_replace} evaluates decoder efficacy under controlled conditions, where all models employ our SegMAN-S Encoder. We benchmark against decoders from representative methods~\cite{xie2021segformer,guo2022segnext,yan2024vwaformer,ni2024cgrseg,yu2024edaformer} with matched model sizes and computational budgets—UPerNet’s channel dimensions were adjusted to ensure equivalent GFLOPs. Our decoder achieves superior mIoU.

Notably, integrating our SegMAN Encoder with third-party decoders yields significant performance gains: +2.1\% for VWFormer, +1.9\% for SegNeXt, and +1.2\% for EDAFormer over their original encoder-decoder configurations (Table~\ref{exp_main_segmentation}). This highlights our encoder’s generalization and plug-and-play capability.



\textbf{Ablation studies on the MMSCopE module.}
 Table~\ref{tab:ablation_within_head_mmscope} presents the impact of various components within the MMSCopE module: (1) Switching to independently scanning each feature map instead of our proposed simultaneous scanning strategy significantly reduces FPS from 128 to 65; (2) The inclusion of multi-scale features yields consistent performance gains. Specifically, removing the feature map at 16x16, 32x32, or 64x64 resolution inside the MMSCopE module results in sub-optimal mIoU, with the removal of the 64x64 feature map resulting in the largest performance drop of 0.7\%. Additionally, using the 64x64 feature map alone without pixel unshuffling and compression leads to a 0.4\% decline.
(3) Channel compression before SS2D proves beneficial, as its removal reduces mIoU by 0.3\%. Alternative channel scaling configuration (1C, 2C, 4C) for the three feature scales yields suboptimal results; (4) The FFN following SS2D provides a marginal benefit of 0.1\%.

\begin{table}[t]
\centering
\scalebox{0.76}{
\begin{tabular}{lccccc}
\toprule
Model Variant & Params (M) & GFLOPs & FPS & mIoU \\
\midrule
\midrule
SegMAN-S & 29.4 & 25.3 & 128 & 51.3 \\
\midrule
\multicolumn{5}{l}{\textit{Scan Strategy}} \\
Independent scans & 28.7 & 25.9 & 65 \textcolor{red}{(-63)} & 51.5 \textcolor{posblue}{(+0.2)} \\
\midrule
\multicolumn{5}{l}{\textit{Multi-scale feature}} \\
w/o 16×16 & 28.9 & 24.8 & 130 \textcolor{posblue}{(+2)} & 51.0 \textcolor{red}{(-0.3)} \\
w/o 32×32 & 28.5 & 24.5 & 132 \textcolor{posblue}{(+4)} & 50.8 \textcolor{red}{(-0.5)} \\
w/o 64×64 & 28.3 & 24.3 & 131 \textcolor{posblue}{(+3)} & 50.6 \textcolor{red}{(-0.7)} \\
64×64 only & 26.2 & 23.6 & 137 \textcolor{posblue}{(+9)} & 50.9 \textcolor{red}{(-0.4)} \\
\midrule
\multicolumn{5}{l}{\textit{Compression}} \\
w/o compression & 30.3 & 26.7 & 98 \textcolor{red}{(-30)} & 51.0 \textcolor{red}{(-0.3)} \\
Channels 1C:2C:4C & 31.2 & 25.3 & 124 \textcolor{red}{(-4)} & 50.8 \textcolor{red}{(-0.5)} \\
\midrule
\multicolumn{5}{l}{\textit{Architecture Components}} \\
w/o FFN & 28.4 & 24.8 & 129 \textcolor{posblue}{(+1)} & 51.2 \textcolor{red}{(-0.1)} \\
\bottomrule
\end{tabular}}
\caption{Ablation studies on the MMSCopE module in SegMAN decoder. FLOPs and FPS are measured at \(512\times 512\) resolution.\vspace{-2mm}}
\label{tab:ablation_within_head_mmscope}
\end{table}

\section{Conclusion}
This work presents a new segmentation network with two novel components: SegMAN Encoder, a hierarchical encoder that synergistically integrates state space models with local attention, and MMSCopE, an efficient decoder module that leverages state space models for multi-scale context extraction.
SegMAN Encoder enables simultaneous modeling of fine-grained spatial details and global contextual information while MMSCopE adaptively scales with the input resolution while processing multi-scale features.
We conduct comprehensive evaluations across multiple benchmarks, which demonstrates that our approach achieves state-of-the-art performance in semantic segmentation.

{
    \small
    \bibliographystyle{ieeenat_fullname}
    \bibliography{main}
}


\clearpage
\setcounter{page}{1}

\newpage
\twocolumn[
    \centering
    \Large
    \textbf{\thetitle} \\
    \vspace{0.5em}
    Supplementary Material \\
    \normalsize
    \vspace{1.5em}
    \normalsize
    \vspace{3.0em}
]

\appendix
\section{Additional ablation studies}
We present additional ablation studies on the MMSCopE module (refer to Figure \ref{method_model} (d) in the main paper) using the ADE20K~\cite{zhou2017ade20k} dataset. Specifically, we investigate the effect of diffusion feature fusion strategies, the effect of feature map resolutions on simultaneous SS2D scans, and then verify the impact of Pixel Shuffle and Pixel Unshuffle operations on preserving feature map details. 

\textbf{Effect of feature fusion in the decoder.}
Table~\ref{tab:ablation_within_head_decoder} illustrates the importance of our feature fusion approach (Figure~\ref{method_model}(c)). Direct prediction from multi-scale context feature \(F'\) results in significant performance degradation (-1.0\%). While incorporating the feature map \(F\) improves performance to 50.9\%, our fusion strategy with stage-specific features \(F'_2\), \(F_{up3}\) \(F_{up4}\) achieves the optimal performance of 51.3\% mIoU.
Removal of average-pooled features and not adding \(F'\) to stage-specific features reduces effectiveness. For a fair comparison, average-pooled features are included in experiments without stage-specific features.

\begin{table}[ht]
\centering
\scalebox{0.8}{
\begin{tabular}{lccc}
\toprule
Input to classifier & Params (M) & GFLOPs  & mIoU \\
\midrule
\midrule
SegMAN-S & 29.4 & 25.3  & 51.3 \\
\midrule
\multicolumn{4}{l}{\textit{w/o stage-specific features}} \\
Concat (\(F\),\(F'\)) & 29.5 & 24.6  & 50.9 \textcolor{red}{(-0.4)} \\
\(F + F'\) & 29.2 & 25.2  & 50.8 \textcolor{red}{(-0.5)} \\
\(F'\) & 29.2 & 24.4  & 50.3 \textcolor{red}{(-1.0)} \\

\midrule
\multicolumn{4}{l}{\textit{with stage-specific features}} \\
w/o avg pool & 29.3 & 25.2  & 50.9 \textcolor{red}{(-0.4)} \\
w/o addition & 29.3 & 25.2  & 51.1 \textcolor{red}{(-0.2)} \\

\bottomrule
\end{tabular}}
\caption{Effect of fusion strategies in the SegMAN decoder.\vspace{-3mm}}
\label{tab:ablation_within_head_decoder}
\end{table}

\textbf{Effect of feature map resolution on SS2D scans.}
We examine how different spatial resolutions for simultaneous SS2D scans influence performance. In our proposed MMSCopE module, feature maps are rescaled and concatenated at a resolution of $\frac{H}{32} \times \frac{W}{32}$ (i.e., $16 \times 16$ for $512 \times 512$ input images in ADE20K). To assess the impact of higher resolutions, we experiment with rescaling feature maps to $\frac{H}{16} \times \frac{W}{16}$ ($32 \times 32$) and $\frac{H}{8} \times \frac{W}{8}$ ($64 \times 64$) resolutions. To gather feature maps at the $\frac{H}{8} \times \frac{W}{8}$ resolution, we upsample the feature maps using the Pixel Shuffle operation~\cite{shi2016real}, which rearranges elements from the channel dimension into the spatial dimension, effectively increasing spatial resolution while preserving feature information. For the $\frac{H}{16} \times \frac{W}{16}$ resolution, we apply Pixel Unshuffle to the $\frac{H}{8} \times \frac{W}{8}$ feature maps, and Pixel Shuffle to the $\frac{H}{32} \times \frac{W}{32}$.

As shown in Table~\ref{sup:additional_ablation_mmscope}, scanning at the proposed $\frac{H}{32} \times \frac{W}{32}$ resolution achieves the best performance. 
Scanning at higher resolutions results in decreased mIoU scores.
This decline occurs because higher resolutions lead to reduced channel dimensions in the feature maps after applying Pixel Shuffle operations. Specifically, the Pixel Shuffle operation decreases the channel dimension by factors of 4 and 16 when upsampling by factors of 2 and 4, respectively. 
This significant reduction in channel dimensions limits the SS2D's learning capacity, thereby negatively impacting performance.

\textbf{Impact of Pixel Unshuffle Operation.} We evaluate replacing the Pixel Unshuffle operation with bilinear interpolation when preparing feature maps for SS2D scanning at the $\frac{H}{32} \times \frac{W}{32}$ resolution. Pixel Unshuffle downscales feature maps without information loss, ensuring the downsampled maps fully represent the original features despite reduced spatial resolution. Processing these maps together enables simultaneous handling of multiple scales, effectively modeling multi-scale information.

As shown in Table~\ref{sup:additional_ablation_mmscope}, substituting Pixel Unshuffle with bilinear interpolation reduces mIoU from 50.0\% to 49.0\%. This confirms that preserving the full representational capacity during downsampling is crucial. Bilinear interpolation, a smoothing operation, loses fine-grained spatial information, leading to diminished segmentation accuracy. Therefore, the Pixel Unshuffle operation is vital for maintaining multi-scale contextual information.

\begin{table}[ht]
\centering
\scalebox{0.8}{
\begin{tabular}{lccc}
\toprule
Model Variant & Params (M) & GFLOPs & mIoU \vspace{1.5mm}\\
\midrule
\midrule
SegMAN-S & 29.9 & 24.6  & 50.0 \\
\midrule
\multicolumn{4}{l}{\textit{Scan Resolution}} \vspace{1.5mm}\\
\(\frac{W}{16}\times  \frac{W}{16}\) (32×32) & 30.5 & 25.2  & 49.5 (-0.5) \vspace{2.5mm}\\
\(\frac{W}{8}\times  \frac{W}{8}\) (64×64) & 29.7 & 26.6  & 49.2 (-0.8) \vspace{1.5mm}\\
\midrule
\multicolumn{4}{l}{\textit{Downsample Method}} \\
Bilinear Interpolation & 29.9 & 24.5 & 49.0 (-1.0)\\
\bottomrule
\end{tabular}}
\caption{Additional ablation studies on the MMSCopE module in SegMAN decoder.\vspace{-2mm}}
\label{sup:additional_ablation_mmscope}
\end{table}

\textbf{Encoder hyperparameter ablations.} Table~\ref{rebuttal_ablations} compares the effect of different window sizes and SS2D parameter settings on SegMAN-T. Default setting (window size [11,9,7,7]) yields best performance.

\begin{table}[ht]
\centering
\scalebox{0.6}{
\begin{tabular}{lcccc} 

\toprule

Encoder config  & Param (M) & GFLOP & ImageNet-1k & ADE20k \\
\midrule
Window size for each stage [13,11,9,7]  & 5.1 \scriptsize{(+0.0)} & 4.8 \scriptsize{(+0.2)} & 76.4 \scriptsize{(+0.2)} & 43.2 \scriptsize{(-0.1)}\\
Window size for each stage [9,7,7,7]  & 5.1 \scriptsize{(+0.0)} & 4.4 \scriptsize{(-0.2)} & 76.1 \scriptsize{(-0.1)} & 43.1 \scriptsize{(-0.2)}\\
SSM expansion ratio \(1\xrightarrow{}2\) & 5.3 \scriptsize{(+0.2)} & 5.2 \scriptsize{(+0.6)} & 76.3 \scriptsize{(+0.1)} & 43.0 \scriptsize{(-0.3)}\\
SSM state dimension N \(1\xrightarrow{}16\) & 5.3 \scriptsize{(+0.2)} & 6.5 \scriptsize{(+1.9)} & 76.5 \scriptsize{(+0.3)} & 43.1 \scriptsize{(-0.2)}\\

\bottomrule
\end{tabular}}
\caption{Effect of Encoder window size and SSM configurations.}
\vspace{-3mm}
\label{rebuttal_ablations}
\end{table}

\section{Detailed backbone comparison}
Table~\ref{appendix:encoder_imagenet_comparison} presents a detailed comparison of ImageNet-1k classification accuracy. Additional representative backbones (PVTv2~\cite{wang2022pvt2}, MaxViT~\cite{tu2022maxvit}, MambaTree~\cite{xiao2025mambatree}) are included for comparison.

\begin{table}[ht]
\setlength{\belowcaptionskip}{-0.2cm}
\centering
\scalebox{0.8}{
\begin{tabular}{lccc} 

\toprule

Models  & Params (M) & GFLOPs & Acc \\
\midrule
\midrule
MiT-B0~\cite{xie2021segformer} & 3.8 &0.60 & 70.5 \\
EFT-T~\cite{yu2024edaformer} &3.7 &0.60 & 72.3 \\ 
MSCAN-T~\cite{guo2022segnext} & 4.2 & 0.89 & 75.9 \\
\rowcolor{ggray} SegMAN-T Encoder & 3.5 & 0.65 & \textbf{76.2} \\
\midrule
MiT-B2~\cite{xie2021segformer} & 24 & 4.0 & 81.6 \\
EFT-B~\cite{yu2024edaformer} & 26 & 4.2 & 82.4 \\
MSCAN-B~\cite{guo2022segnext} & 27 & 4.4 & 83.0 \\
Swin-T~\cite{liu2021swin} & 28 & 4.5 & 81.2\\
PVTv2-B2~\cite{wang2022pvt2} & 25 & 4.0 & 79.8 \\ 
ConvNeXt-T~\cite{liu2022convnext} & 28 & 4.5 & 82.1 \\
InternImage-T~\cite{wang2023internimage} & 30 & 5.0 & 83.5 \\
MaxViT-T~\cite{tu2022maxvit} & 31 & 5.6 & 83.7 \\
ViM-S~\cite{zhu2024vim} & 26 & - & 81.6 \\
VMamba-T~\cite{liu2024vmamba} & 29 & 4.9 & 82.6 \\
MambaTreev-T~\cite{xiao2025mambatree} & 30 & 4.8 & 83.4 \\
SparX-Mamba-T~\cite{lou2024sparx} & 27 & 5.2 & 83.5 \\
\rowcolor{ggray} SegMAN-S Encoder & 26 & 4.1 & \textbf{84.0} \\
\midrule
MiT-B3~\cite{xie2021segformer} & 45 & 6.9 & 83.1 \\
MSCAN-L~\cite{guo2022segnext} & 45 & 9.1 & 83.9 \\
Swin-S~\cite{liu2021swin} & 50 & 8.7 & 83.2\\
PVTv2-B3~\cite{wang2022pvt2} & 45 & 6.9 & 83.2 \\
ConvNeXt-S~\cite{liu2022convnext} & 50 & 8.7 & 83.1 \\
InternImage-S~\cite{wang2023internimage} & 50 & 8.0 & 84.2 \\
MaxViT-S~\cite{tu2022maxvit} & 69 & 11.7 & 84.5 \\
VMamba-S~\cite{liu2024vmamba} & 50 & 8.7 & 83.6 \\
MambaTreeV-S~\cite{xiao2025mambatree} & 51 & 8.5 & 84.2 \\
SparX-Mamba-S~\cite{lou2024sparx} & 47 & 9.3 & 84.2 \\
\rowcolor{ggray} SegMAN-B Encoder & 45 & 9.9 & \textbf{85.1} \\
\midrule
MiT-B5~\cite{xie2021segformer} & 82 & 11.8 & 83.8 \\
Swin-B~\cite{liu2021swin} & 88 & 15.4 & 83.5\\
PVTv2-B5~\cite{wang2022pvt2} & 82 & 11.8 & 83.8 \\
ConvNeXt-B~\cite{liu2022convnext} & 89 & 15.4 & 83.8 \\
InternImage-B~\cite{wang2023internimage} & 97 & 16.7 & 84.9 \\
MaxViT-B~\cite{tu2022maxvit} & 120 & 24.0 & 84.9 \\
ViM-B~\cite{zhu2024vim} & 98 & - & 83.2 \\
VMamba-B~\cite{liu2024vmamba} & 89 & 15.9 & 84.5 \\
MambaTreeV-B~\cite{xiao2025mambatree} & 91 & 15.1 & 84.8 \\
SparX-Mamba-B~\cite{lou2024sparx} & 84 & 15.9 & 84.5 \\
\rowcolor{ggray} SegMAN-L Encoder & 81 & 16.8 & \textbf{85.5} \\
\bottomrule
\end{tabular}}
\caption{Detailed comparison of classification accuracy and computational complexity (FLOPs at \(224\times 224\) resolution) of encoder architectures on ImageNet-1K.}
\label{appendix:encoder_imagenet_comparison}
\end{table}

\definecolor{myred}{RGB}{200, 0, 0}

\section{Generalization}
Table~\ref{rebuttal_generalization} empirically demonstrates the modular compatibility of SegMAN components across two representative frameworks: SegNeXt and CGRSeg. Replacing SegNeXt’s encoder with our SegMAN-S Encoder reduces parameters by 9\% and GFLOPs by 15\% while improving ADE20K mIoU by +0.7\%; substituting its decoder achieves +0.4\% mIoU at 14\% lower computation. Similarly, integrating our encoder into CGRSeg yields +1.7\% mIoU, while our decoder enhances its performance by +0.8\% mIoU. These results quantify the efficacy of our encoder and decoder in balancing accuracy-efficiency trade-offs, validating that either component can independently upgrade existing pipelines. The bidirectional improvements underscore SegMAN’s plug-and-play adaptability, where each module achieves an optimal balance of performance gains (up to +1.7\% mIoU) and computational pragmatism across diverse architectures.

\begin{table}[ht]
\centering
\scalebox{0.6}{
\begin{tabular}{lllccc} 

\toprule

Configuration & Feature Encoder & Decoder & Param & GFLOP &  ADE20k \\
\midrule
SegNeXt  & MSCAN-B & HAM & 27.7 & 34.9 & 48.5 \\
\rowcolor{ggray} SegNeXt + our encoder & SegMAN-S & HAM & 25.3 & 29.5  & \textbf{49.2} \scriptsize{(\color{myred}{+0.7}}) \\
\rowcolor{ggray} SegNeXt + our decoder & MSCAN-B & Ours & 30.6 & 29.9 & \textbf{48.9} \scriptsize{(\color{myred}{+0.4}}) \\
\midrule
CGRSeg & EfficientFormerV2-L & CGRHead & 35.7 & 16.5 & 47.3 \\
\rowcolor{ggray} CGRSeg + our encoder &  SegMAN-S & CGRHead & 44.2 & 24.5 & \textbf{49.0} \scriptsize{(\color{myred}{+1.7}}) \\
\rowcolor{ggray} CGRSeg + our decoder  & EfficientFormerV2-L & Ours & 29.9 & 17.9 & \textbf{48.1} \scriptsize{(\color{myred}{+0.8}}) \\

\bottomrule
\end{tabular}}
\caption{Encoder and Decoder generalization results.}
\vspace{-2.5mm}
\label{rebuttal_generalization}
\end{table}

\section{Panoptic and instance segmentation}
To demonstrate task-agnostic capabilities, we deploy our SegMAN-S Encoder in Mask DINO~\cite{li2023maskDINO} for panoptic and instance segmentation. Replacing its default ResNet50 backbone with our ImageNet-1k pretrained encoder as well as the MiT-B2~\cite{xie2021segformer} backbone in SegFormer. We maintain Mask DINO's architecture while increasing batch size from 16 to 48 for training efficiency.

As shown in Table~\ref{rebuttal_mask_dino}, SegMAN-S achieves 49.6 instance AP (+3.3 over ResNet50, +2.0 over MiT-B2) and 56.8 panoptic PQ (+3.8/+2.1) while operating at 283 GFLOPs, which is 6\% fewer than ResNet50 (286 GFLOPs) and 10\% fewer than MiT-B2 (315 GFLOPs). Despite comparable parameter counts (48.3M vs. 48.5M MiT-B2), our encoder delivers superior multi-task performance, validating its effectiveness beyond semantic segmentation.

\begin{table}[ht]
\centering
\scalebox{0.6}{
\begin{tabular}{lccccc} 

\toprule

 Encoder & Param (M) & GFLOP & Instance AP & Panoptic PQ \\
\midrule
 ResNet50 & 52 & 286 & 46.3 & 53.0  \\
 MiT-B2 & 48.5 & 315 & 47.6 &  54.7 \\
\rowcolor{ggray} SegMAN-S Encoder & 48.3 & 283 & \textbf{49.6} \scriptsize{(\color{myred}{+3.3}})  & \textbf{56.8} \scriptsize{(\color{myred}{+3.8}})  \\

\bottomrule
\end{tabular}}
\vspace{-2.5mm}
\caption{Panoptic and instance segmentation using Mask DINO.}
\label{rebuttal_mask_dino}
\end{table}

\section{Qualitative examples}
\begin{figure*}[t]
\centering
\includegraphics[width=0.98\textwidth]{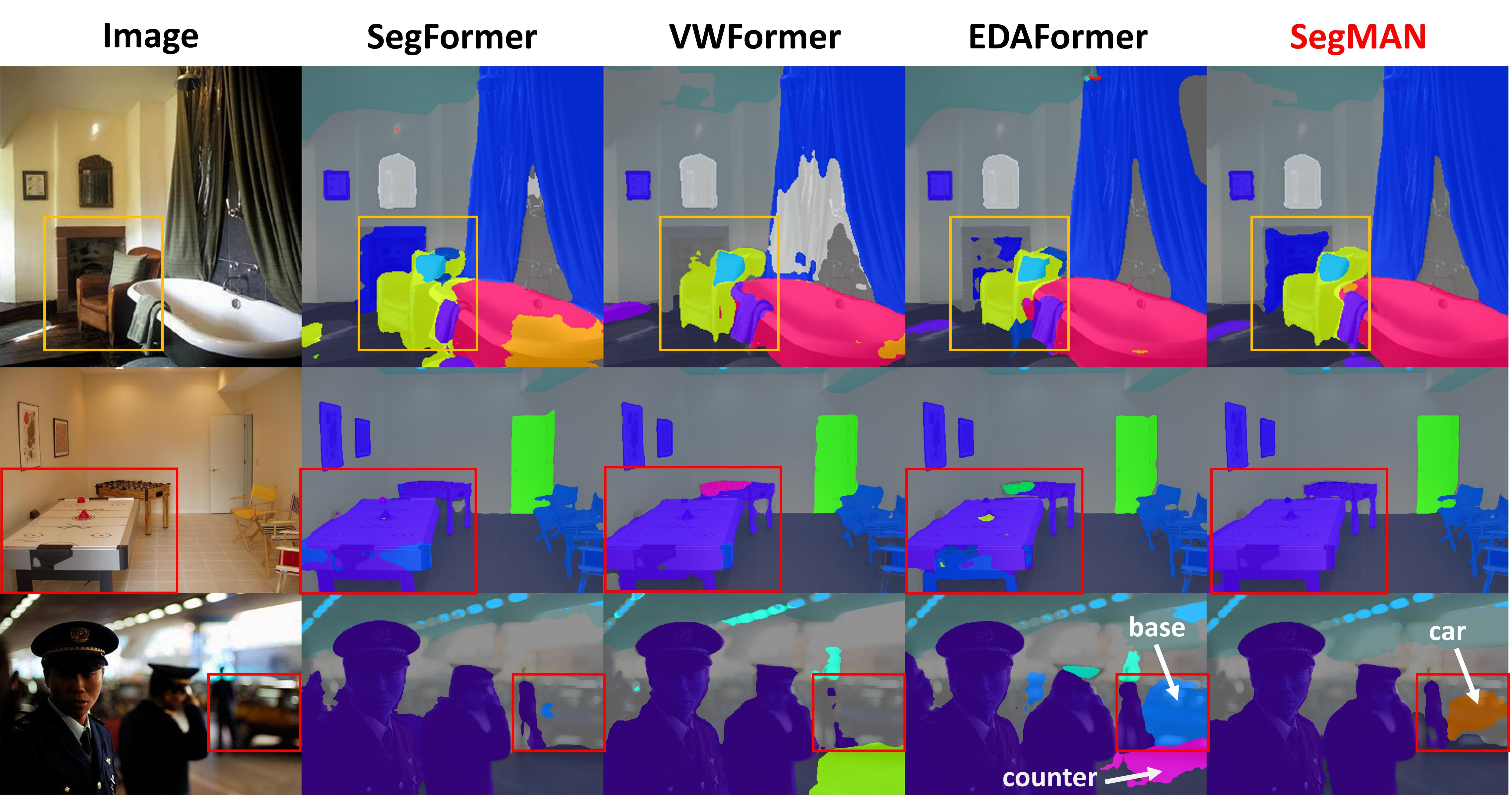}
\caption{\textbf{Qualitative results on ADE20K.} Zoom in for best view.}
\label{sup_ade_comparison}
\end{figure*}

\begin{figure*}[t]
\centering
\includegraphics[width=0.98\textwidth]{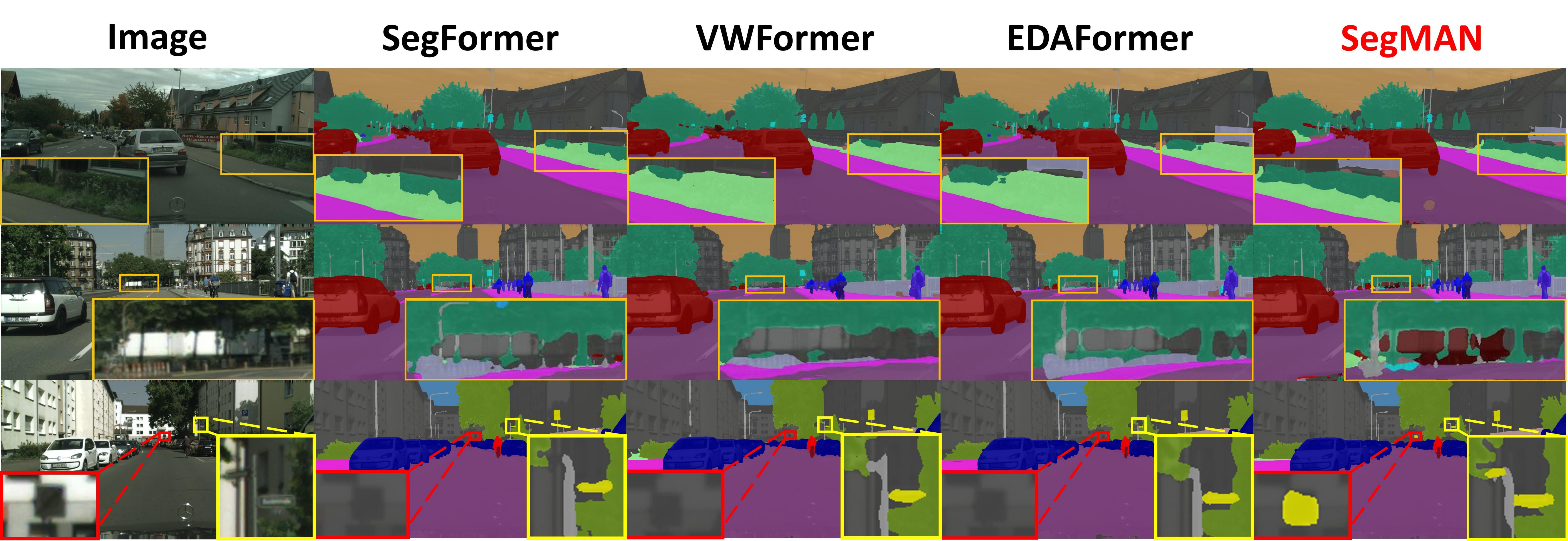}
\caption{\textbf{Qualitative results on Cityscapes.} Zoom in for best view.}
\label{sup_city_comparison}
\end{figure*}

\begin{figure*}
\centering
\includegraphics[width=0.9\textwidth]{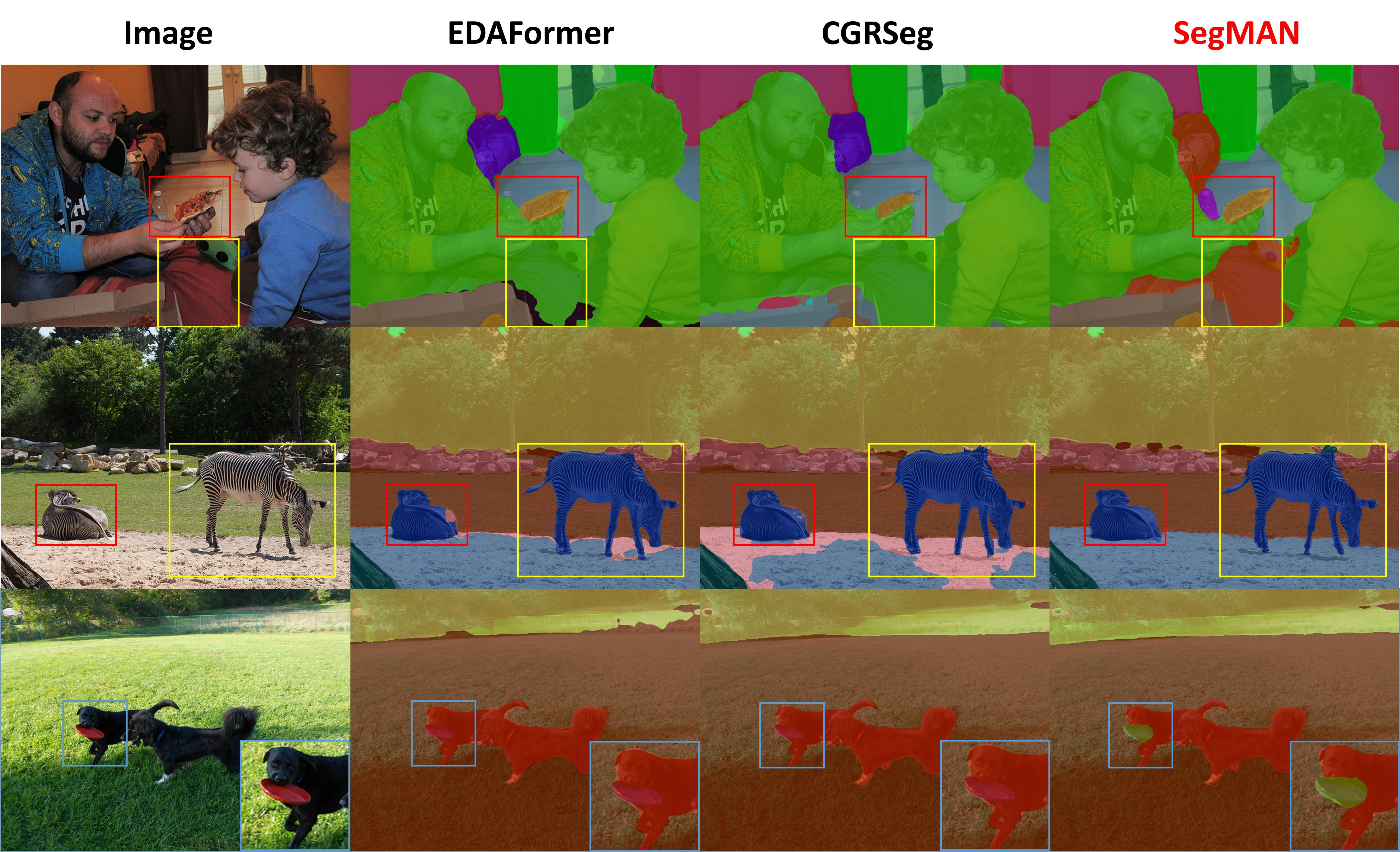}
\caption{\textbf{Qualitative results on COCO-Stuff-164K.} We do not compare with SegFormer as its COCO-Stuff checkpoints are not released. Zoom in for best view.}
\label{sup_coco_comparison}
\end{figure*}

We present qualitative examples of SegMAN's segmentation results on ADE20K Figures~\ref{sup_ade_comparison}, Cityscapes~\ref{sup_city_comparison}, and COCO-Stuff-164K~\ref{sup_coco_comparison}. For COCO-Stuff, comparisons are made with VWFormer and EDAFormer only, since the checkpoints for other segmentation models are not released. These figures illustrate SegMAN's capability to capture both fine-grained local dependencies and long-range contextual information. Compared to other segmentation methods, SegMAN yields more precise boundaries and accurately identifies intricate details within the scenes.
These qualitative results verify our quantitative findings, highlighting the benefits of SegMAN's ability to capture fine-grained details while maintaining global context, which is unattainable by existing approaches.

\end{document}